\documentclass[letterpaper, 10 pt, conference]{ieeeconf}  

\IEEEoverridecommandlockouts

\usepackage{graphics} 
\usepackage{mathptmx} 
\usepackage{amsmath} 
\usepackage{amssymb}  
\usepackage{subcaption}

\title{\LARGE \bf
Data Augmentation for 3DMM-based Arousal-Valence Prediction for HRI
}

\author{Christian Arzate Cruz*$^{1}$ Yotam Sechayk*$^{2}$ Takeo Igarashi$^{2}$ and Randy Gomez$^{1}$
\thanks{* Joint first authorship.}
\thanks{$^{1}$ Honda Research Institute Japan (HRI-JP),
        Wako City, Japan.
        {\tt\small christian.arzate@jp.honda-ri.com}}%
\thanks{$^{2}$ The University of Tokyo (UTokyo), Tokyo, Japan.
        {\tt\small sechayk-yotam@g.ecc.u-tokyo.ac.jp}}%
}

\usepackage{todonotes}
\usepackage[normalem]{ulem}

\IEEEoverridecommandlockouts
\newcommand\copyrighttext{%
  \footnotesize \textcopyright 2024 IEEE. Personal use of this material is permitted. Permission from IEEE must be obtained for all other uses, in any current or future media, including reprinting/republishing this material for advertising or promotional purposes, creating new collective works, for resale or redistribution to servers or lists, or reuse of any copyrighted component of this work in other works.}
\newcommand\copyrightnotice{%
\begin{tikzpicture}[remember picture,overlay]
\node[anchor=south,yshift=10pt] at (current page.south) {\fbox{\parbox{\dimexpr\textwidth-\fboxsep-\fboxrule\relax}{\copyrighttext}}};
\end{tikzpicture}%
}

\IEEEaftertitletext{\vspace{-2\baselineskip}}

\begin{document}

\maketitle
\copyrightnotice
\thispagestyle{empty}
\pagestyle{empty}

\begin{abstract}
    Humans use multiple communication channels to interact with each other. For instance, body gestures or facial expressions are commonly used to convey an intent. The use of such non-verbal cues has motivated the development of prediction models. One such approach is predicting arousal and valence (AV) from facial expressions. However, making these models accurate for human-robot interaction (HRI) settings is challenging as it requires handling multiple subjects, challenging conditions, and a wide range of facial expressions. In this paper, we propose a data augmentation (DA) technique to improve the performance of AV predictors using 3D morphable models (3DMM). We then utilize this approach in an HRI setting with a mediator robot and a group of three humans. Our augmentation method creates synthetic sequences for underrepresented values in the AV space of the SEWA dataset, which is the most comprehensive dataset with continuous AV labels. Results show that using our DA method improves the accuracy and robustness of AV prediction in real-time applications. The accuracy of our models on the SEWA dataset is $0.793$ for arousal and valence.
\end{abstract}

\section{Introduction}
Robots that interact with people can benefit by detecting non-verbal cues in communication since they often provide a better look at the internal state of humans \cite{Burgoon2016nonverbal, Pentland2010honest}. In particular, the human-robot interaction (HRI) community has special attention to finding novel techniques to identify and use non-verbal cues such as facial expressions \cite{Arakawa2018, Lin2020review}. One of the main challenges in this area is to predict the emotional state of humans in real-time.
In this paper, we propose a novel augmentation approach to improve a model that predicts the arousal and valance (AV) of humans in a way that overcomes the main shortcomings in our previous HRI experiments \cite{Cooper2023}.

Our testbed, \textit{The Talking Room}, consists of a multi-group conversation where young teen participants share their life experiences with our robot -- Haru. The main goal of this group activity is to support pro-social skills that connects high-school students of different backgrounds. Haru is the mediator of these conversations. We designed this group activity to foster rich group interactions such as turn-taking, and attention changes. In our previous work, we found that predicting the AV of the participants is key to understanding social acknowledgment and social engagement \cite{Cooper2023}. Besides, other works in the HRI community have shown that arousal and valence are two fundamental cues to the quality of the interaction \cite{Gervasi2023, Abrams2020, Anwar2016, Budman1993}.

The level of arousal represents the intensity of the emotion, while valence is how pleasant or unpleasant the emotion is \cite{Russell1980}. However, current approaches to predict AV from facial expressions have room to improve in HRI settings, such as The Talking Room. In particular, our AV model design adopts three main requirements: (1) real-time prediction in a group setting, (2) robustness to different lighting conditions, (3) invariance to head orientation and position, and (4) high accuracy in both positive and negative emotions.

To achieve requirements (1) through (3), our model uses EMOCA \cite{Danvevcek2022}, which is a  3D morphable model (3DMM) fitting pipeline. The pipeline is used to extract the facial expression features. By using 2D videos as input, we can extract vectors of coefficients that represent the facial expression temporally. 3DMM techniques are robust to different lighting conditions, head positions, and maintain temporal coherency \cite{Tellamekala2023}. Additionally, using a fast face detector such as BlazeFace \cite{Bazarevsky2019} allows us to achieve real-time performance.

\begin{figure}[t]
    \vspace{0.5cm} 
    \centering
    \includegraphics[width=0.85\linewidth]{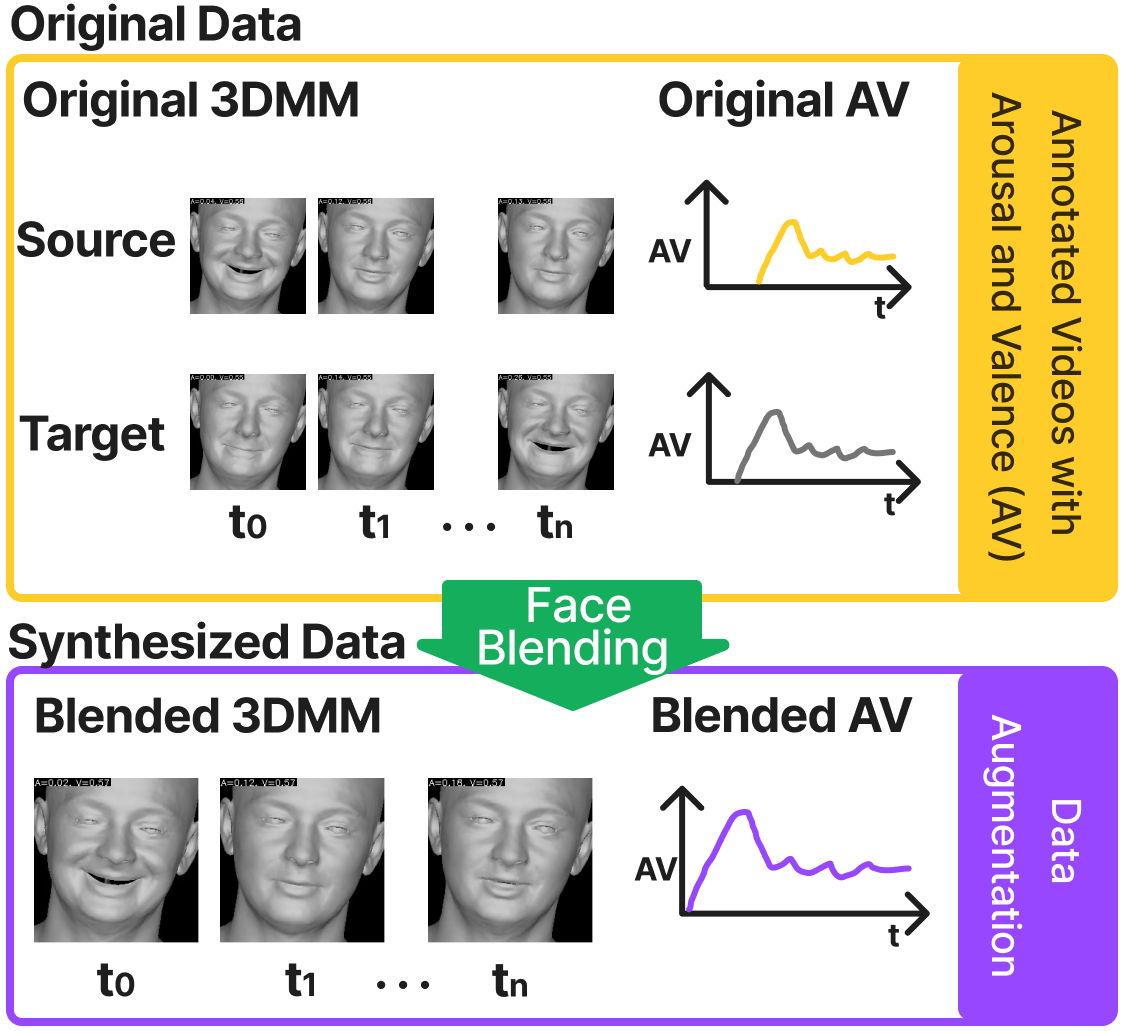}
    \caption{Overview of our 3DMM-based data augmentation method.}
    \label{fig:overview}
\end{figure}

For our requirement (4), we propose a novel data augmentation method that creates synthetic 3DMMs for underrepresented values in the AV space of the SEWA dataset \cite{Kossaifi2019} (we present an overview of our method in Figure \ref{fig:overview}). The main idea is to create new sequences of facial expressions by blending the coefficients of videos that are in the same area of the AV space. We can achieve face blending by using the coefficients of the 3DMM fitting. We use the coefficients of the 3DMM fitting to create a sequence of facial expressions by finding other videos that generate a consistent result. To the best of our knowledge, our work is the first to use a 3DMM-based model to create ``synthetic videos'' of facial expressions.

We explore several methods to generate synthetic sequences of facial expressions. As well as heuristics on how to select the source and target videos for blending. For labels, we use a weighted average of the arousal and valence values respectively, between the source and the target. Where the weight is derived from the blending method used. The results are smooth expression transitions localized between the source and target videos.

Finally, we performed a study to evaluate the effect of our 3DMM-based data augmentation techniques on the performance of our model in the SEWA dataset \cite{Kossaifi2019}. Then, we evaluated the performance of the baseline and the best-performing model in The Talking Room. Our results show that by using our AV predictor model trained with our proposed data augmentation method, we can get better accuracy and robustness in an HRI group setting. These results are encouraging since they show that our model can be used in real-time HRI applications.

\section{Related Work}
This section is an overview of research on facial expression recognition and the use of data augmentation techniques in this field. Besides, we introduce the most common 3D morphable models (3DMM) approaches.

\subsection{3D Morphable Models of Faces}
A 3D morphable model (3DMM) is a statistical model that represents the expression and shape of a face \cite{Blanz2003}. The parameters of a particular face are a set of coefficients that modify a generic model of the face to fit a target face. Some of the first methods for 3DMMs are the Basel Face Model (BFM) \cite{Paysan2009} and the Surrey Face Model (SFM) \cite{Huber2016}. More recently, the 3DMMs have improved in terms of fidelity and accuracy \cite{Chang2018, Li2017, Feng2021}.

For the EMOCA model \cite{Danvevcek2022}, their authors incorporate the use of an emotion recognition model to improve the quality of the 3DMM fitting. That is, the EMOCA model incorporates in its loss function the perceptual emotion of faces. In this manner, the EMOCA model can fit the 3DMM to a face in a way that the emotion of the face is preserved. For this reason, we use this model to extract the features of the face in our approach.

\subsection{Facial Expression Recognition (FER)}
The standard in expression analysis is the use of 2D features from videos \cite{Kollias2017, Sanchez2018, Toisoul2021, Kollias2018}. The main idea is to extract landmarks from the face and use them as input for a model. Arguably, the most common landmarks are the ones introduced in the 300-W dataset \cite{Sagonas2016}. These landmarks are salient points in the face (e.g., mouth, eyes, or nose). In contrast, the use of a 3DMM finds the parameters of a 3D model that best fits a given face. The output of the 3DMM is a set of coefficients that represent the facial expression, head position, and shape. By using an 3DMM, we skip the step of extracting the landmarks directly from the RGB videos.

Most of the previous works have used 3DMMs to extract expression features and use them to train models that predict categorical emotions \cite{Wen2003, Bejaoui2017, Savran2008}. According to our exhaustive research, the first work that used a 3DMM to predict arousal and valence was \cite{Tellamekala2023}. They extract the facial features using the 3DMM fitting called EMOCA \cite{Danvevcek2022} and train a model to predict arousal and valence. Our work is based on this approach, but we focus on improving the performance of the model by adapting it to overcome the limitations in our HRI setting.

\subsection{Data Augmentation Techniques for FER}
To get an accurate facial expression recognition model, we need to train it with a large dataset that contains a wide variety of facial expressions, and a large number of subjects. These requirements are necessary since different people can express the same emotion differently, and cultural differences can affect the way people express their emotions \cite{Kossaifi2019}. One way to get an accurate model with a smaller dataset is to use data augmentation (DA) techniques.

In the context of FER, DA methods are classified in data warping and oversampling augmentations \cite{Shorten2019}. Data warping techniques generate new images by applying linear geometric transformations (e.g., rotation, scaling, flipping). On the other hand, oversampling techniques create synthetic data for specific target values (e.g., using a GAN to generate new images of a specific emotion \cite{Kammoun2022}). For a survey on the use of DA techniques in FER, we refer the reader to \cite{Porcu2020}.

The use of a 3DMM model in our approach allows us to skip the use of DA to improve the quality of the feature detection since EMOCA is already robust enough to provide good results in different situations. However, we can still use DA techniques to generate synthetic sequences of emotional expressions to expand the original training set. In this paper, we explore several DA techniques that are suitable for 3DMM-based models. We evaluate the effect of these techniques on the performance of our model in the SEWA dataset \cite{Kossaifi2019}.

\subsection{AI Models in Mixed Groups}
Our understanding of the dynamics of communication between a robot and a group of humans (a mixed group) is still limited \cite{Jung2018robots}. What we know is that when people interact with a robot in a group, they express their emotions differently compared to when they are alone with a robot \cite{Leite2015comparing}. As a consequence, the performance of AI models that predict emotions from facial expressions can be affected by the group setting. To try to overcome this issue, we design our AV predictor model to be robust in an HRI setting.

\section{Pipeline Overview}
\label{sec:overview}
\begin{figure*}[th]
    \centering
    \includegraphics[width=0.95\textwidth]{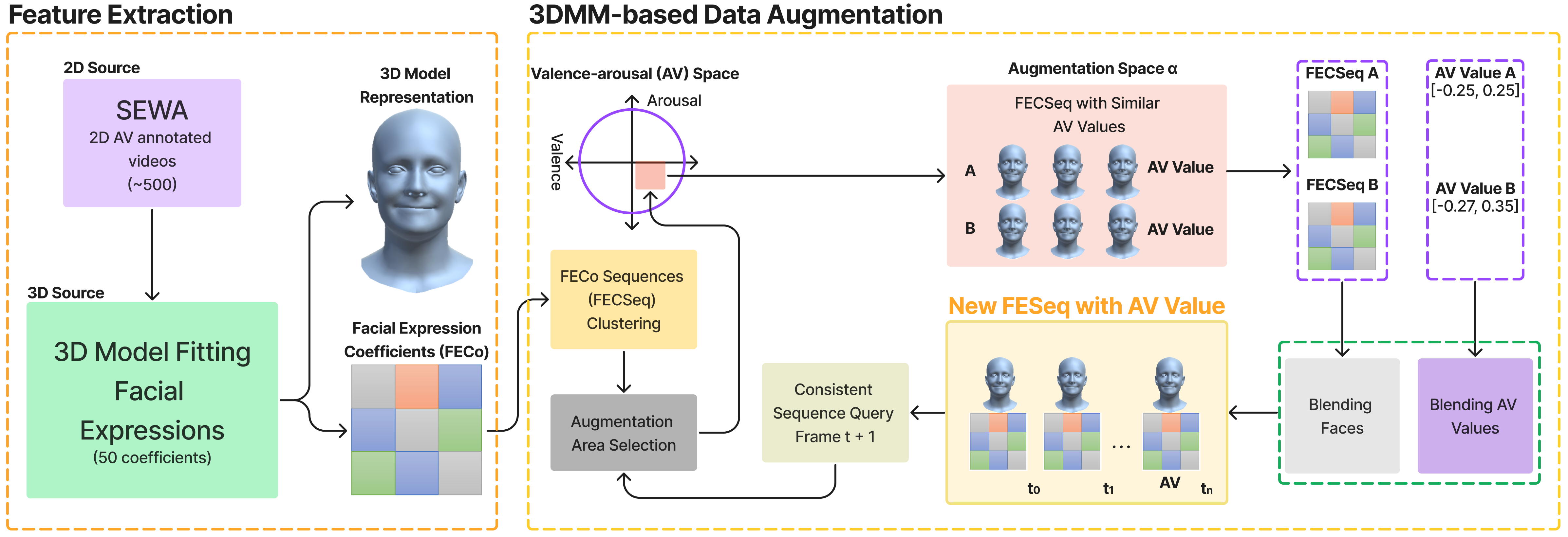}
    \caption{An overview of our pipeline to generate a model with a DA-enhanced dataset.}
    \label{fig:model}
\end{figure*}

We present an overview of our pipeline, the requirements and problem statement, data handling, and the architecture of the model used. In Figure \ref{fig:model}, we present all the components pipeline to create our arousal and valence prediction model.

\subsection{Requirements in The Talking Room}
Based on previous pilot studies in our testbed, we identified the following requirements for our AV predictor model. First, the model needs to perform in real-time for online human-robot communication. Additionally, the model needs to be robust to different lighting conditions, head orientations, and head and body positions. Finally, the model needs to have high accuracy in both positive and negative emotions to correctly represent the participants' emotional states. These requirements define the problem we wish to address and guide the design of our model and the augmentation technique we propose.

\subsection{Data Handling}
For data handling, we perform a sample-independent split of the dataset. We use $80\%$ of the data for training, $10\%$ for validation, and $10\%$ for test. For the DA-enhanced model, we use the same train/validation split as the baseline with the synthetic sequences added to the training set. We keep the original label values in the range of $[-1, 1]$ for each frame in the dataset.

\subsection{Model Architecture}
\label{sec:model}
Our model is based on the work by \cite{Tellamekala2023} to make the comparisons easier. The model consists of a bi-directional Gated Recurrent Unit (GRU) network with two layers of size $128$. The input of the model is a sequence of $100$ frames (two seconds). The output is a tuple of two values that represent the arousal and valence of the sequence.

For \textbf{training}, we use a batch size of $4$ and dropout in the GRU layers ($0.5$). The learning rate is $0.001$, and the model is trained for $\sim 25$ epochs. We also use the Adam \textbf{optimizer} with parameters for its weight decay and $T_0$ of $0.0001$ and $20$, respectively. Besides, it includes a CosineAnnealing scheduler with warm restarts \cite{Loshchilov2017}.

We implemented the \textbf{loss function} presented in \cite{Kossaifi2020}. The loss function is a combination of the Root Mean Squared Error (RMSE), the Pearson product-moment Correlation Coefficient, and Lin's Concordance Correlation Coefficient.

\subsection{Baseline Model}
For evaluation, we used a baseline model following prior work \cite{Tellamekala2023}. The baseline model uses the same architecture and parameters as well as the same data split.

\section{Feature Extraction}
\label{sec:feature_extraction}
In this section, we present the dataset and the expression extraction method. We use the EMOCA model \cite{Danvevcek2022} to extract the expression features for each frame in the SEWA dataset \cite{Kossaifi2019}. Figure \ref{fig:model} shows an overview of this process (left side).

\subsection{Face Detection}
\label{sec:face_detection}
The input to EMOCA is a cropped face, which makes the face detection (FD) algorithm impact our model's performance in two ways. First, the accuracy and speed of the FD vary considerably. And some even fail to find faces in specific lighting conditions (e.g., dark scenes). Second, the bounding boxes of the faces differ depending on the method used; this affects the quality of EMOCA's output. For example, if the bounding box crops out part of the chin, the predicted expression will differ from the input. After performing tests with multiple off-the-shelf FD, we choose to use BlazeFace (BF) \cite{Bazarevsky2019}. BF strikes a balance between speed and compatibility.

\subsection{Dataset}
\label{sec:dataset}
For our work, we need a video-based dataset (\textbf{2D source}) with continuous annotations of arousal and valence. We use the SEWA dataset \cite{Kossaifi2019} since it is the biggest dataset that meets these requirements. The SEWA dataset is publicly available and contains $538$ labeled videos of $348$ subjects with a diverse cultural background. Besides, it is gender balanced, with age range between $18$ and $65$. The videos' frame rate is $50$ frames per second. The dataset contains annotations of arousal and valence for each frame of the videos. These values are the consensus of six different expert annotators. The annotations are continuous values in the range $[-1, 1]$. We chose to use the SEWA dataset since it is the biggest dataset for affect estimation on videos.

\subsection{Facial Expression Coefficients}
Since our model is designed using 3DMM in mind, we process the SEWA dataset \cite{Kossaifi2019} using EMOCA \cite{Danvevcek2022}. Similar to \cite{Tellamekala2023}, we take as input a 3D image (i.e., single frame), detect the face and generate a crop. Then, the output is several vectors of coefficients that represent: the facial expression, head position and orientation, shape, and texture. In this paper, we focus exclusively on the facial expression coefficients (FECo). The FECo is represented by a $50$ dimensional vector, $\mathbf{F}(t) \in \mathbb{R}^{50}$ with values ranging from $-4$ to $4$. Some outliers exist, mostly in situations where the face is partially occluded.


\section{3DMM-based Data Augmentation}
\label{sec:augmentation}
One of the key requirements for The Talking Room demo is a model that is accurate in both positive and negative emotions (i.e., robustness). However, the SEWA dataset is unbalanced in terms of the distribution of the arousal and valence values (see Figure \ref{fig:av_comparison}). To address this issue, we propose data augmentation (DA) techniques that are suitable for 3DMM-based models.

In this section, we introduce the 3DMM-based DA technique that we use for our training dataset. Figure \ref{fig:model} shows our proposed DA flow, suitable for 3DMM-based models. The augmentation can be separated into (1) source video selection, (2) target video selection, and (3) blending. In the following sections, we present the details of each step.

\subsection{Source Video Selection}
The first step in our DA process is to select source videos for blending. Initially, we identify the areas in the AV space that are underrepresented. The concept is to create synthetic videos that strengthen the model's robustness to different expressions. To achieve this, we present two AV space approaches, frame-based and video-based. Each approach has a different effect on the blended output and the clustering methodology. We aim to achieve an AV space that is more balanced and representative of in-the-wild emotions.

\subsubsection{AV Space Clustering}
\paragraph{Frame-based}
Frame-based blending refers to blending done per frame. I.e., for each frame of the source video we locate the most similar frame (based on AV values) from the AV-clustering space. We then blend the two frames to create a new frame which will be the frame of the augmented sample. This means that the clustering is done on AV values per frame. As the nature of AV labeling is not entirely expression-dependent, two frames with similar AV values could have widely different expressions. Therefore, this approach generates unrealistic synthetic videos where the expression transition is not smooth.


\paragraph{Video-based}
Video-based blending refers to blending done per video. I.e., for the source video, we locate the most similar video (based on mean AV values) from the AV-clustering space. We then blend the two videos to create a new video which will be the augmented sample. This means that the clustering is done on mean AV values per video. This approach maintains a smooth expression transition. We used this approach for our final model.

\subsubsection{Source Video Selection}
To achieve our goal, we locate clusters of AV values with less representation. We explore two methods to achieve this, the first uses KMeans clustering and the second uses discrete K-bin clustering.

\paragraph{KMeans Clustering}
We employ the KMeans clustering algorithm, using the mean AV values of each video as input. We then select videos from clusters with the lowest number of videos, in a greedy manner.

\paragraph{Discrete K-bin Clustering}
We divide the AV space into discrete bins based on the mean AV values of each video. This means that we divide the arousal and valence to $\sqrt{K}$ bins, resulting in $K$ bins. We then select videos from bins with the lowest number of videos, in a greedy manner.

\subsection{Target Video Selection}
The second step in our DA process is to select target videos for blending. Target videos are selected per source video. We explore three selection methods, random, near, and similar. Our experiments show that the similar selection method is the most effective for our use case.

\subsubsection{Random Selection}
Random selection refers to selecting a random video from the dataset. This method is the most straightforward, yet does not result in a balancing effect.

\subsubsection{Near Selection}
Near selection refers to selecting a video that is near the source video. We define near as belonging to the same cluster as the source video. This method results in a balancing effect, yet the synthetic sequences can be unnatural.

\subsubsection{Similar Selection}
Similar selection refers to selecting a video that is similar to the source video in 3DMM feature space. We define similarity as the video where the distance between the mean and variance from the source and the target will be minimized. This method results in a balancing effect with more natural sequences. We used this method for our final model.

\begin{figure}[th]
    \centering
    \begin{subfigure}[b]{0.45\textwidth}
        \centering
        \includegraphics[width=\textwidth]{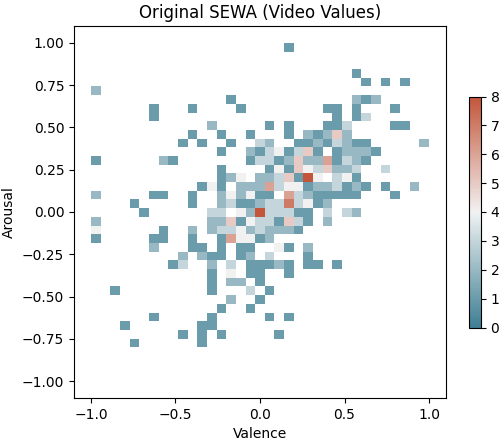}
    \end{subfigure}

    \begin{subfigure}[b]{0.45\textwidth}
        \centering
        \includegraphics[width=\textwidth]{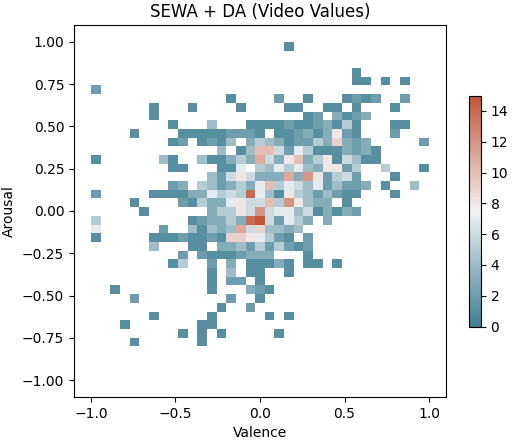}
    \end{subfigure}
    \caption{Comparison of arousal and valence distributions.}
    \label{fig:av_comparison}
\end{figure}

\subsection{Blending}
The third and final step in our DA process is to blend the facial expression coefficients and AV labels.

\subsubsection{Expression Blending}
We start by explaining how to blend the facial expression coefficients. We explored several blending techniques for synthetic sequence creation: random blending, selective weighted blending, and full weighted blending. Our experiments show that the full weighted blending technique is the most effective.

\paragraph{Random Blending}
Random blending is done by fixing a random subset of the coefficients from the source video and replacing the remaining coefficients with the coefficients of the target video.

\paragraph{Selective Weighted Blending}
Selective weighted blending is done by fixing a random subset of the coefficients from the source video and performing a weighted blend of the remaining coefficients with the coefficients of the target video. The weight is a random value between $0.25$ and $0.75$.

\paragraph{Full Weighted Blending}
Full weighted blending is done by performing a weighted blend of all the coefficients of the source video with the coefficients of the target video. The weight is a random value between $0.25$ and $0.75$. This technique is the most effective in our experiments.

\begin{figure}
    \centering
    \includegraphics[width=0.95\columnwidth]{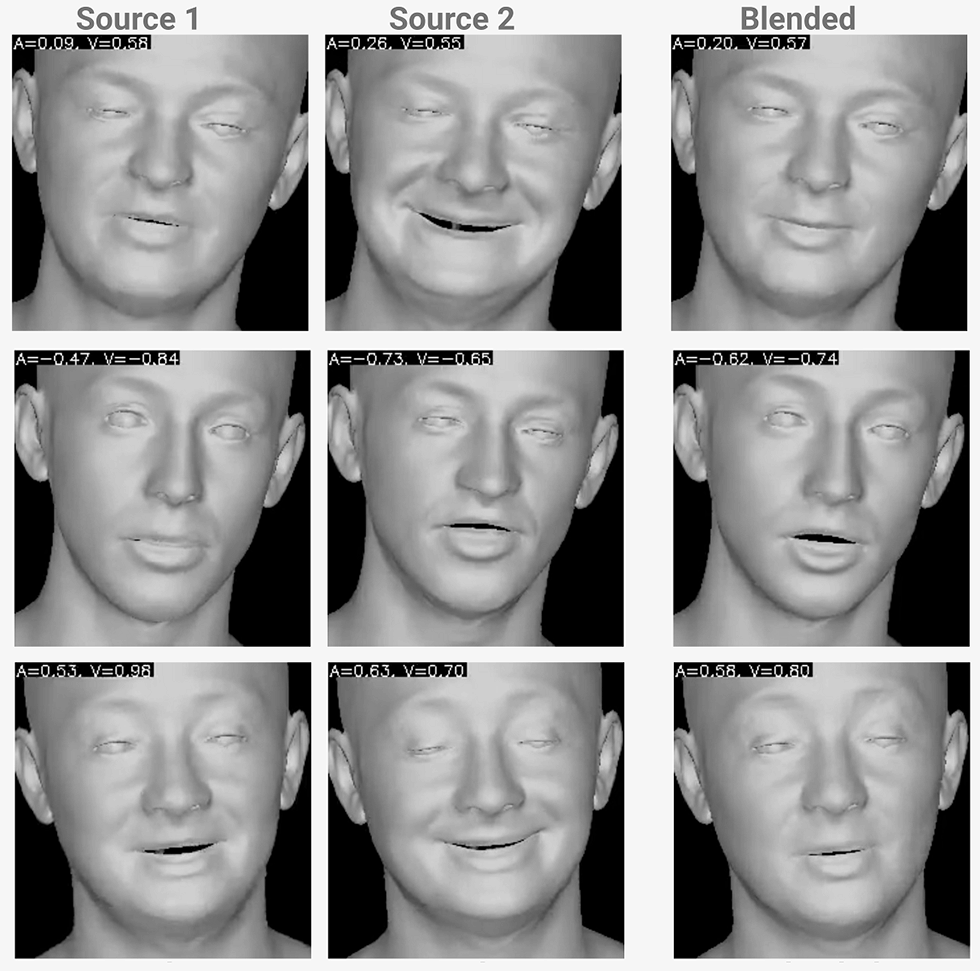}
    \caption{Examples of synthetic sequences generated using our approach.}
    \label{fig:blend}

\end{figure}

We present examples of the best blending technique in Figure \ref{fig:blend}.

\subsubsection{Arousal and Valence Blending}
Finally, arousal and valence blending refers to computing the labels for the synthetic frames. We perform this process by computing the weighted average of the arousal and valence values of the source and target videos, per frame. This uses the same weight defined in the expression blending step.

\section{Synthetic Dataset}

As a result of our 3DMM-based DA techniques, we created an augmented dataset that is more balanced in terms of the distribution of the arousal and valence values. We generated 600 sequences, using 100 source videos and 6 target videos for each, with an average length of 586 frames. We present the results of our ablation study in the next section.

As we can see in Figure \ref{fig:av_comparison}, our DA techniques have a positive effect on the distribution of the arousal and valence values in the SEWA dataset.

\section{SEWA Evaluation}
\label{sec:results}
In this section, we present the accuracy of our AV predictor model in the SEWA dataset \cite{Kossaifi2019}. As \textbf{evaluation metric}, we use the Lin's Concordance Correlation Coefficient (CCC) \cite{Lawrence1989}. We chose this metric to be consistent with the rest of the literature. We compute the CCC with the next formula:

\begin{equation*}
    CCC = \frac{2  \sigma_x \sigma_y PCC(x, y)}{\sigma_x^2 + \sigma_y^2 + (\mu_x - \mu_y)^2}
\end{equation*}

where $PCC(x, y)$ is the Pearson product-moment Correlation Coefficient of the two variables, $\sigma_x$ and $\sigma_y$ are the standard deviations of the two variables, and $\mu_x$ and $\mu_y$ are the means of the two variables.

\begin{table}[ht]
    \centering
    \caption{Arousal and valence prediction accuracy on the SEWA dataset.}
    \label{tab:ccc}
    \begin{tabular}{lccc}
        \hline
        \textbf{Model}                           & \textbf{Arousal} & \textbf{Valence} & \textbf{Mean}    \\
        ~                                        & \textbf{CCC}     & \textbf{CCC}     & \textbf{CCC}     \\
        \hline
        Mitenkova et al. \cite{Mitenkova2019}    & $0.392$          & $0.469$          & $0.430$          \\
        Toisoul et al. \cite{Toisoul2021}        & $0.610$          & $0.650$          & $0.630$          \\
        Kossaifi et al. \cite{Kossaifi2019}      & $0.520$          & $0.750$          & $0.635$          \\
        Sanchez et al. \cite{Sanchez2021}        & $0.640$          & $0.750$          & $0.695$          \\
        Tellamekala et at. \cite{Tellamekala2023} & $0.716$          & $0.775$          & $0.745$          \\
        \textbf{Our model}                       & $\textbf{0.799}$ & $\textbf{0.788}$ & $\textbf{0.793}$ \\
        \hline
    \end{tabular}
\end{table}

As we can see in Table \ref{tab:ccc}, our model has a higher accuracy compared to the rest of the state-of-the-art models tested on the SEWA dataset. In the next section, we present an evaluation of our model in a realistic HRI setting.

\section{The Talking Room Evaluation}
\label{sec:demo}
\begin{figure*}[!t]
    \centering
    \includegraphics[width=0.95\textwidth]{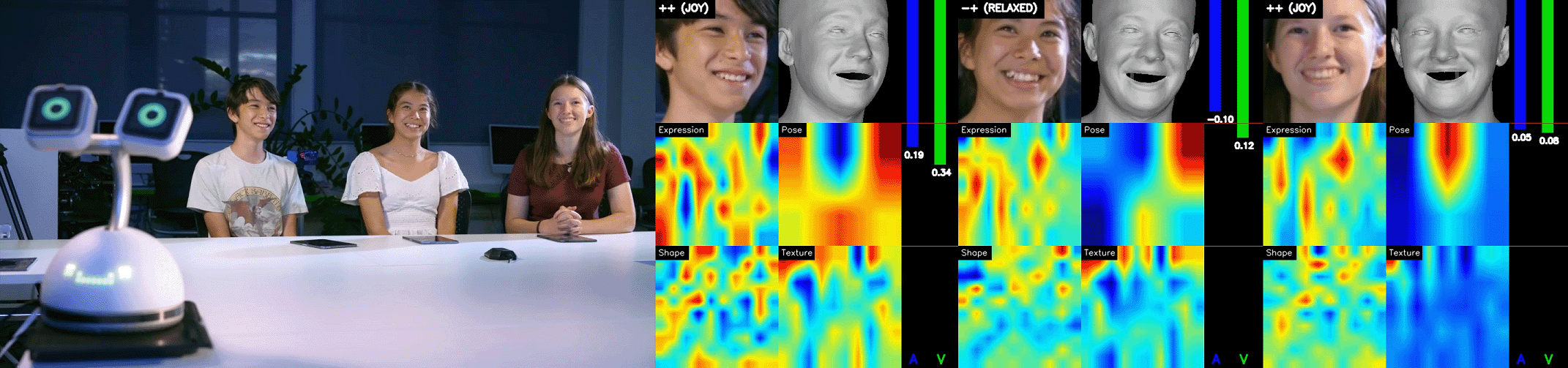}
    \caption{Arousal and valence prediction accuracy in The Talking Room.}
    \label{fig:tsc_av}
\end{figure*}

Our study, The Talking Room (see Figure \ref{fig:tsc_av}), consists of two groups of three children in two different locations. Each group has a Haru robot and a screen to communicate with the other group \cite{Cooper2023}. Haru has the role of a mediator in the conversation. The topics of this scenario involve sharing life experiences related to school (e.g., favorite class, etc.).

\subsection{Quantitative Analysis}

To be able to evaluate our AV predictor model in our HRI setting, we implemented an application that works in real-time. We use an RGB camera running at $30$ fps to capture the video of the participants. In Figure \ref{fig:tsc_av}, we show the interface of our application that shows the arousal and valence of the participants. We also present the 3D representation of their facial expression, and 2D representations of the expression, and position coefficients from EMOCA \cite{Danvevcek2022}.

We used two computer setups to evaluate the performance. Both use an Intel i9 processor, with one equipped with an \verb|RTX 3090-Ti| and the other \verb|RTX 3090|. For the former, our pipeline using the BlazeFace detector, runs at $15$fps. And the latter on the same detector runs as $12$fps. We also performed a detailed analysis measuring the time (ms) each part of our pipeline takes to execute. Our results indicate that the BlazeFace takes $27.1$ms per frame, the inference time of EMOCA is $47.6$ms, and the inference of our AV model is $0.7$ms. In Table \ref{tab:times} we present a time analysis of our pipeline using the \verb|RTX 3090| GPU (time is in milliseconds and averaged over a $7$ sec video with one face).

\begin{table}[!tbh]
    \centering
    \caption{Time analysis of our pipeline.}
    \label{tab:times}
    \begin{tabular}{lccccc}
        \hline
        \textbf{FD}        & \textbf{Detect} & \textbf{Crop} & \textbf{EMOCA} & \textbf{AV Model} & \textbf{Total} \\
        \hline
        SFD                & 44.4            & 5.8           & 55.9           & 0.8               & 110.9          \\
        \textbf{BlazeFace} & 27.1            & 5.5           & 47.6           & 0.7               & 84.4           \\
        HOG                & 31.6            & 6.3           & 47.6           & 0.8               & 90.3           \\
        CNN                & 532.3           & 5.1           & 44.2           & 0.8               & 586.5          \\
        DNN                & 11.1            & 5.6           & 45.7           & 0.8               & 67.7           \\
        \hline
    \end{tabular}
\end{table}

\subsection{Quality Evaluation}

As a test to verify how the AV values of the group change while talking with our robot, we measure the mean arousal and valence per frame in the groups. This strengthens our evaluation by testing our model in a different setting and a user's demographics not present in SEWA (children).

\begin{figure}[tb]
    \centering
    \begin{subfigure}[b]{0.47\textwidth}
        \centering
        \includegraphics[width=\textwidth]{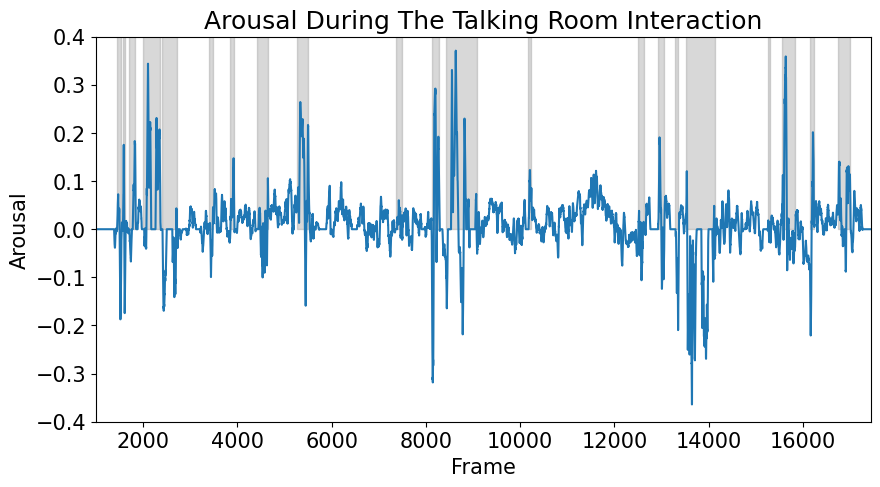}
    \end{subfigure}
    \hfill
    \begin{subfigure}[b]{0.47\textwidth}
        \centering
        \includegraphics[width=\textwidth]{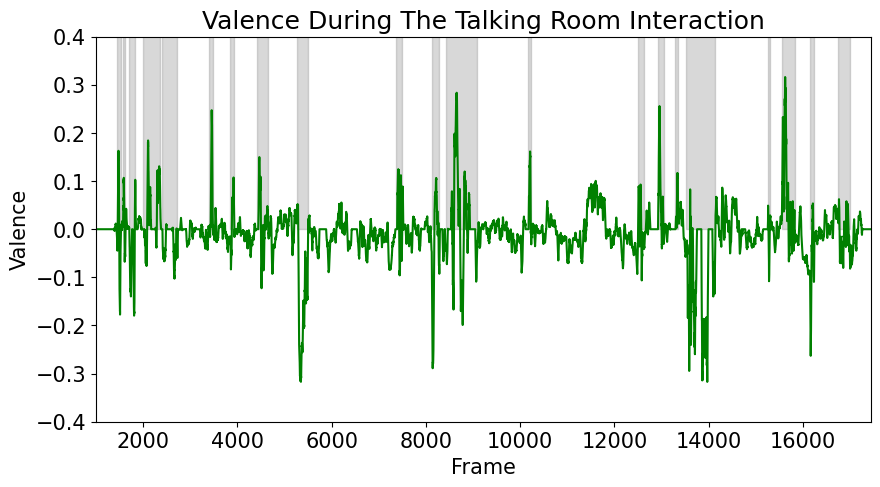}
    \end{subfigure}
    \caption{Arousal and valence over time.}
    \label{fig:av_time}
\end{figure}

To quantify the performance of our model in the HRI setting, we verify the changes in AV values of the group change while talking with our robot, Haru. In particular, we measure the mean arousal and valence per frame in the groups. We choose this metric since it is a good indicator of the quality of the interaction. We present the AV values per frame in Figure \ref{fig:av_time}. As we can see, both the mean arousal and valence of the groups present bigger changes when the children interact with the robot (we highlight these moments in the figure). Besides, the change in the valence of the emotions and their intensity is consistent with our previous studies that were manually evaluated by a behavioral expert \cite{Cooper2023}. For instance, close to the end of the interaction children seem unhappy while Haru explains that the activity is about to end.

Most of the emotional exchanges between the children and Haru are positive in both arousal and valence. Furthermore, our model was able to capture negative changes in the AV values when the children were talking about subjects that they did not seem to like. This validates the accuracy of our model to predict both positive and negative emotions in a realistic HRI setting.

\begin{figure}[tbh]
    \centering
    \includegraphics[width=0.9\columnwidth]{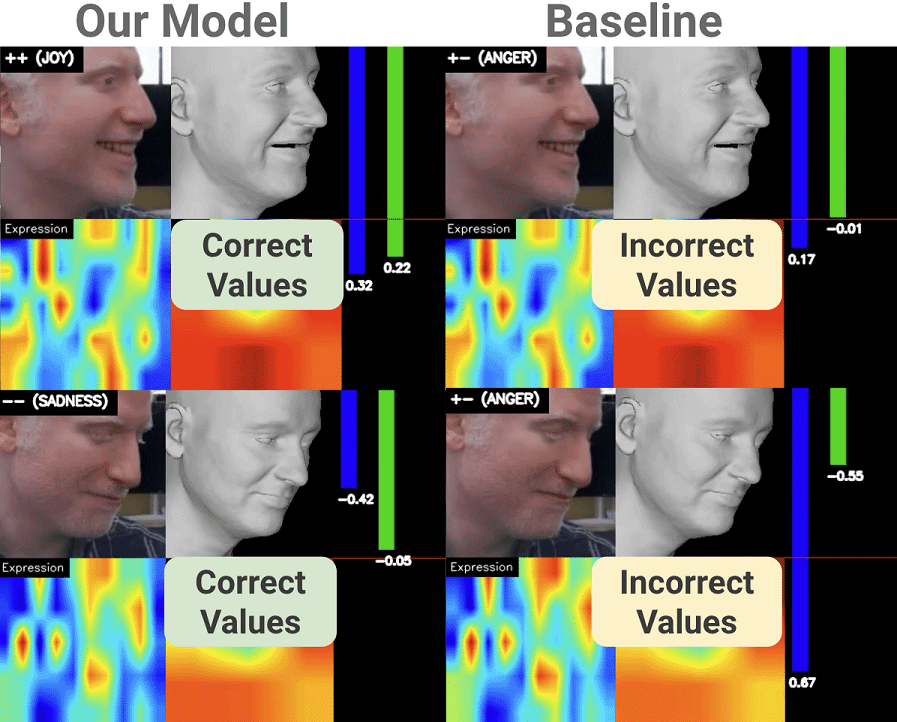}
    \caption{AV prediction examples.}
    \label{fig:far_left}
\end{figure}

Another advantage we found while using our model is that it is robust in an HRI setting to different head positions. In particular, we found that our model can handle well when people look to their far left or right. We present examples of this advantage in Figure \ref{fig:far_left}.

\section{Limitations and Future Work}
Study limitations include the small number of participants (N = $6$). We plan to address this limitation with larger user studies in the future. We also plan to evaluate the performance of our model in other HRI settings. In particular, we are interested in exploring the use of our model in settings that involve a wider range of emotions from the participants.

Another limitation is that our model is not able to predict the emotions of the participants when they are looking down or their face is partially occluded. We can mitigate the impact of these situations by implementing a tracking system that uses previous AV predictions when the faces of participants are not fully visible.

Finally, we plan to explore the use of our model to predict more complex social features that are relevant to The Talking Room. For instance, we are interested in exploring the use of our model as a basis to predict the level of social acknowledgment and social engagement of the participants. We believe that our model can be used to provide a deeper insight into the socio-emotional interaction in the HRI. Additionally, we are interested in exploring various additional DA techniques to further improve the AV prediction model.

\section{Conclusions}
We presented a novel AV predictor model that is designed to be robust in an HRI setting. We used a 3DMM fitting model to extract the facial expression features. We also proposed a novel DA method that creates synthetic sequences for underrepresented values in the AV space. By combining the benefits of using a 3DMM-based model and our DA method, we created an augmented version of the SEWA dataset \cite{Kossaifi2019} that is more balanced in terms of the distribution of arousal and valence values. We used our dataset to train our AV predictor which achieved state-of-the-art results. Furthermore, we evaluated the performance of our model in an HRI scenario with three children and a robot. Our results show that using our AV predictor is robust enough to provide insight into the socio-emotional interactions. These results are encouraging since they show that our model can be used in real-time HRI applications.

\bibliographystyle{IEEEtran}
\bibliography{Bibliography}

\begin{thebibliography}{10}
\providecommand{\url}[1]{#1}
\csname url@samestyle\endcsname
\providecommand{\newblock}{\relax}
\providecommand{\bibinfo}[2]{#2}
\providecommand{\BIBentrySTDinterwordspacing}{\spaceskip=0pt\relax}
\providecommand{\BIBentryALTinterwordstretchfactor}{4}
\providecommand{\BIBentryALTinterwordspacing}{\spaceskip=\fontdimen2\font plus
\BIBentryALTinterwordstretchfactor\fontdimen3\font minus \fontdimen4\font\relax}
\providecommand{\BIBforeignlanguage}[2]{{%
\expandafter\ifx\csname l@#1\endcsname\relax
\typeout{** WARNING: IEEEtran.bst: No hyphenation pattern has been}%
\typeout{** loaded for the language `#1'. Using the pattern for}%
\typeout{** the default language instead.}%
\else
\language=\csname l@#1\endcsname
\fi
#2}}
\providecommand{\BIBdecl}{\relax}
\BIBdecl

\bibitem{Burgoon2016nonverbal}
J.~K. Burgoon, V.~Manusov, and L.~K. Guerrero, \emph{Nonverbal communication}.\hskip 1em plus 0.5em minus 0.4em\relax Routledge, 2016.

\bibitem{Pentland2010honest}
A.~Pentland, \emph{Honest signals: how they shape our world}.\hskip 1em plus 0.5em minus 0.4em\relax MIT press, 2010.

\bibitem{Arakawa2018}
R.~Arakawa, S.~Kobayashi, Y.~Unno, Y.~Tsuboi, and S.-i. Maeda, ``Dqn-tamer: Human-in-the-loop reinforcement learning with intractable feedback,'' \emph{arXiv preprint arXiv:1810.11748}, 2018.

\bibitem{Lin2020review}
J.~Lin, Z.~Ma, R.~Gomez, K.~Nakamura, B.~He, and G.~Li, ``A review on interactive reinforcement learning from human social feedback,'' \emph{IEEE Access}, vol.~8, pp. 120\,757--120\,765, 2020.

\bibitem{Cooper2023}
S.~Cooper and R.~Gomez, ``Towards the use of a social mediator robot in a school setting,'' in \emph{GROUND workshop}, 2023.

\bibitem{Gervasi2023}
R.~Gervasi, L.~Mastrogiacomo, and F.~Franceschini, ``An experimental focus on learning effect and interaction quality in human--robot collaboration,'' \emph{Production Engineering}, pp. 1--26, 2023.

\bibitem{Abrams2020}
A.~M. Abrams and A.~M. R.-v. der P{\"u}tten, ``I--c--e framework: Concepts for group dynamics research in human-robot interaction: Revisiting theory from social psychology on ingroup identification (i), cohesion (c) and entitativity (e),'' \emph{International Journal of Social Robotics}, vol.~12, pp. 1213--1229, 2020.

\bibitem{Anwar2016}
K.~Anwar, ``Working with group-tasks and group cohesiveness.'' \emph{International Education Studies}, vol.~9, no.~8, pp. 105--111, 2016.

\bibitem{Budman1993}
S.~H. Budman, S.~Soldz, A.~Demby, M.~Davis, and J.~Merry, ``What is cohesiveness? an empirical examination,'' \emph{Small Group Research}, vol.~24, no.~2, pp. 199--216, 1993.

\bibitem{Russell1980}
J.~A. Russell, ``A circumplex model of affect.'' \emph{Journal of personality and social psychology}, vol.~39, no.~6, p. 1161, 1980.

\bibitem{Danvevcek2022}
R.~Dan{\v{e}}{\v{c}}ek, M.~J. Black, and T.~Bolkart, ``Emoca: Emotion driven monocular face capture and animation,'' in \emph{Proceedings of the IEEE/CVF Conference on Computer Vision and Pattern Recognition}, 2022, pp. 20\,311--20\,322.

\bibitem{Tellamekala2023}
M.~K. Tellamekala, {\"O}.~S{\"u}mer, B.~W. Schuller, E.~Andr{\'e}, T.~Giesbrecht, and M.~Valstar, ``Are 3d face shapes expressive enough for recognising continuous emotions and action unit intensities?'' \emph{IEEE Transactions on Affective Computing}, 2023.

\bibitem{Bazarevsky2019}
V.~Bazarevsky, Y.~Kartynnik, A.~Vakunov, K.~Raveendran, and M.~Grundmann, ``Blazeface: Sub-millisecond neural face detection on mobile gpus,'' \emph{arXiv preprint arXiv:1907.05047}, 2019.

\bibitem{Kossaifi2019}
J.~Kossaifi, R.~Walecki, Y.~Panagakis, J.~Shen, M.~Schmitt, F.~Ringeval, J.~Han, V.~Pandit, A.~Toisoul, B.~Schuller \emph{et~al.}, ``Sewa db: A rich database for audio-visual emotion and sentiment research in the wild,'' \emph{IEEE transactions on pattern analysis and machine intelligence}, vol.~43, no.~3, pp. 1022--1040, 2019.

\bibitem{Blanz2003}
V.~Blanz and T.~Vetter, ``Face recognition based on fitting a 3d morphable model,'' \emph{IEEE Transactions on pattern analysis and machine intelligence}, vol.~25, no.~9, pp. 1063--1074, 2003.

\bibitem{Paysan2009}
P.~Paysan, R.~Knothe, B.~Amberg, S.~Romdhani, and T.~Vetter, ``A 3d face model for pose and illumination invariant face recognition,'' in \emph{2009 sixth IEEE international conference on advanced video and signal based surveillance}.\hskip 1em plus 0.5em minus 0.4em\relax Ieee, 2009, pp. 296--301.

\bibitem{Huber2016}
P.~Huber, G.~Hu, R.~Tena, P.~Mortazavian, W.~P. Koppen, W.~J. Christmas, M.~R{\"a}tsch, and J.~Kittler, ``A multiresolution 3d morphable face model and fitting framework,'' in \emph{International conference on computer vision theory and applications}, vol.~5.\hskip 1em plus 0.5em minus 0.4em\relax SciTePress, 2016, pp. 79--86.

\bibitem{Chang2018}
F.-J. Chang, A.~T. Tran, T.~Hassner, I.~Masi, R.~Nevatia, and G.~Medioni, ``Expnet: Landmark-free, deep, 3d facial expressions,'' in \emph{2018 13th IEEE International Conference on Automatic Face \& Gesture Recognition (FG 2018)}.\hskip 1em plus 0.5em minus 0.4em\relax IEEE, 2018, pp. 122--129.

\bibitem{Li2017}
T.~Li, T.~Bolkart, M.~J. Black, H.~Li, and J.~Romero, ``Learning a model of facial shape and expression from 4d scans,'' vol.~36, no.~6, 2017.

\bibitem{Feng2021}
Y.~Feng, H.~Feng, M.~J. Black, and T.~Bolkart, ``Learning an animatable detailed 3d face model from in-the-wild images,'' \emph{ACM Transactions on Graphics (ToG)}, vol.~40, no.~4, pp. 1--13, 2021.

\bibitem{Kollias2017}
D.~Kollias, M.~A. Nicolaou, I.~Kotsia, G.~Zhao, and S.~Zafeiriou, ``Recognition of affect in the wild using deep neural networks,'' in \emph{Proceedings of the IEEE Conference on Computer Vision and Pattern Recognition Workshops}, 2017, pp. 26--33.

\bibitem{Sanchez2018}
E.~Sanchez, G.~Tzimiropoulos, and M.~Valstar, ``Joint action unit localisation and intensity estimation through heatmap regression,'' \emph{arXiv preprint arXiv:1805.03487}, 2018.

\bibitem{Toisoul2021}
A.~Toisoul, J.~Kossaifi, A.~Bulat, G.~Tzimiropoulos, and M.~Pantic, ``Estimation of continuous valence and arousal levels from faces in naturalistic conditions,'' \emph{Nature Machine Intelligence}, vol.~3, no.~1, pp. 42--50, 2021.

\bibitem{Kollias2018}
D.~Kollias and S.~Zafeiriou, ``Aff-wild2: Extending the aff-wild database for affect recognition,'' \emph{arXiv preprint arXiv:1811.07770}, 2018.

\bibitem{Sagonas2016}
C.~Sagonas, E.~Antonakos, G.~Tzimiropoulos, S.~Zafeiriou, and M.~Pantic, ``300 faces in-the-wild challenge: Database and results,'' \emph{Image and vision computing}, vol.~47, pp. 3--18, 2016.

\bibitem{Wen2003}
Z.~Wen \emph{et~al.}, ``Capturing subtle facial motions in 3d face tracking,'' in \emph{Proceedings Ninth IEEE International Conference on Computer Vision}.\hskip 1em plus 0.5em minus 0.4em\relax IEEE, 2003, pp. 1343--1350.

\bibitem{Bejaoui2017}
H.~Bejaoui, H.~Ghazouani, and W.~Barhoumi, ``Fully automated facial expression recognition using 3d morphable model and mesh-local binary pattern,'' in \emph{Advanced Concepts for Intelligent Vision Systems: 18th International Conference, ACIVS 2017, Antwerp, Belgium, September 18-21, 2017, Proceedings 18}.\hskip 1em plus 0.5em minus 0.4em\relax Springer, 2017, pp. 39--50.

\bibitem{Savran2008}
A.~Savran, N.~Aly{\"u}z, H.~Dibeklio{\u{g}}lu, O.~{\c{C}}eliktutan, B.~G{\"o}kberk, B.~Sankur, and L.~Akarun, ``Bosphorus database for 3d face analysis,'' in \emph{Biometrics and Identity Management: First European Workshop, BIOID 2008, Roskilde, Denmark, May 7-9, 2008. Revised Selected Papers 1}.\hskip 1em plus 0.5em minus 0.4em\relax Springer, 2008, pp. 47--56.

\bibitem{Shorten2019}
C.~Shorten and T.~M. Khoshgoftaar, ``A survey on image data augmentation for deep learning,'' \emph{Journal of big data}, vol.~6, no.~1, pp. 1--48, 2019.

\bibitem{Kammoun2022}
A.~Kammoun, R.~Slama, H.~Tabia, T.~Ouni, and M.~Abid, ``Generative adversarial networks for face generation: A survey,'' \emph{ACM Computing Surveys}, vol.~55, no.~5, pp. 1--37, 2022.

\bibitem{Porcu2020}
S.~Porcu, A.~Floris, and L.~Atzori, ``Evaluation of data augmentation techniques for facial expression recognition systems,'' \emph{Electronics}, vol.~9, no.~11, p. 1892, 2020.

\bibitem{Jung2018robots}
M.~Jung and P.~Hinds, ``Robots in the wild: A time for more robust theories of human-robot interaction,'' pp. 1--5, 2018.

\bibitem{Leite2015comparing}
I.~Leite, M.~McCoy, D.~Ullman, N.~Salomons, and B.~Scassellati, ``Comparing models of disengagement in individual and group interactions,'' in \emph{Proceedings of the Tenth Annual ACM/IEEE International Conference on Human-Robot Interaction}, 2015, pp. 99--105.

\bibitem{Loshchilov2017}
\BIBentryALTinterwordspacing
I.~Loshchilov and F.~Hutter, ``{SGDR:} stochastic gradient descent with warm restarts,'' in \emph{5th International Conference on Learning Representations, {ICLR} 2017, Toulon, France, April 24-26, 2017, Conference Track Proceedings}.\hskip 1em plus 0.5em minus 0.4em\relax OpenReview.net, 2017. [Online]. Available: \url{https://openreview.net/forum?id=Skq89Scxx}
\BIBentrySTDinterwordspacing

\bibitem{Kossaifi2020}
J.~Kossaifi, A.~Toisoul, A.~Bulat, Y.~Panagakis, T.~M. Hospedales, and M.~Pantic, ``Factorized higher-order cnns with an application to spatio-temporal emotion estimation,'' in \emph{Proceedings of the IEEE/CVF Conference on Computer Vision and Pattern Recognition}, 2020, pp. 6060--6069.

\bibitem{Lawrence1989}
I.~Lawrence and K.~Lin, ``A concordance correlation coefficient to evaluate reproducibility,'' \emph{Biometrics}, pp. 255--268, 1989.

\bibitem{Mitenkova2019}
A.~Mitenkova, J.~Kossaifi, Y.~Panagakis, and M.~Pantic, ``Valence and arousal estimation in-the-wild with tensor methods,'' in \emph{2019 14th IEEE International Conference on Automatic Face \& Gesture Recognition (FG 2019)}.\hskip 1em plus 0.5em minus 0.4em\relax IEEE, 2019, pp. 1--7.

\bibitem{Sanchez2021}
E.~Sanchez, M.~K. Tellamekala, M.~Valstar, and G.~Tzimiropoulos, ``Affective processes: stochastic modelling of temporal context for emotion and facial expression recognition,'' in \emph{Proceedings of the IEEE/CVF Conference on Computer Vision and Pattern Recognition}, 2021, pp. 9074--9084.

\end{thebibliography}
\end{document}